\documentclass[runningheads,a4paper]{llncs}

\usepackage{amssymb}
\usepackage{graphicx}
\usepackage{subcaption}
\usepackage{multirow}
\usepackage{geometry}
\usepackage{nicefrac}
\usepackage{ctable}
\usepackage[symbol]{footmisc}
\usepackage{booktabs,tabularx}
\usepackage{floatrow}
\newcolumntype{x}{>{\raggedright\arraybackslash}X}
\usepackage{bbding}
\usepackage{amsmath}
\usepackage[export]{adjustbox}
\usepackage[colorlinks=true,
            linkcolor=magenta,
            urlcolor=gray,
            citecolor=blue]{hyperref}\usepackage{slashbox}
\usepackage{arydshln}
\usepackage{floatrow}
\newfloatcommand{capbtabbox}{table}[][\FBwidth]
\usepackage{blindtext}

\newcolumntype{t}{>{\hsize=0.25\hsize}x}
\newcolumntype{s}{>{\hsize=0.8\hsize}x}
\newcolumntype{s}{>{\hsize=0.75\hsize}x}
\newcolumntype{X}{>{\hsize=1.5\hsize}x}

\usepackage{url}
\urldef{\mailsa}\path|anjany.sekuboyina@tum.de|
\urldef{\mailsb}\path||

\newcommand{\siunit}[2]{$#1\,\mathrm{#2}$}

\begin{document}

\mainmatter  

\title{Btrfly Net: Vertebrae Labelling with Energy-\\based Adversarial Learning of Local Spine Prior\footnote[2]{Published as conference paper in Medical Image Computing and Computer Assisted Intervention -- MICCAI 2018}}

\titlerunning{Energy-based adversarial learning for vertebrae labelling}

%
%
\newcommand*\samethanks[1][\value{footnote}]{\footnotemark[#1]}
\author{Anjany Sekuboyina$^{1,2}$ \and Markus Rempfler$^{1}$  \and Jan Kuka\v{c}ka$^{1,2}$ \and Giles Tetteh$^{1}$  \and \\Alexander Valentinitsch$^{2}$ \and Jan S. Kirschke$^{2,}$\thanks{Joint supervising authors.} \and \\Bjoern H. Menze$^{1,}$\samethanks}
\authorrunning{Sekuboyina et al.}
\institute{$^{1}$Department of Informatics, Technical University of Munich, Germany\\ $^{2}$Department of Neuroradiology, Klinikum rechts der Isar, Germany\\
\mailsa}
\toctitle{Energy-based adversarial learning for vertebrae labelling}
\maketitle

\begin{abstract}
Robust localisation and identification of vertebrae is essential for automated spine analysis. The contribution of this work to the task is two-fold: (1) Inspired by the human expert, we hypothesise that a sagittal and coronal reformation of the spine contain sufficient information for labelling the vertebrae. Thereby, we propose a butterfly-shaped network architecture (termed Btrfly Net) that efficiently combines the information across reformations. (2) Underpinning the Btrfly net, we present an energy-based adversarial training regime that encodes local spine structure as an anatomical prior into the network, thereby enabling it to achieve state-of-art performance in all standard metrics on a benchmark dataset of 302 scans without any post-processing during inference.
\end{abstract}

\section{Introduction}
The localisation and identification of anatomical structures is a significant part of any medical image analysis routine. In spine's context, labelling of vertebrae has immediate diagnostic and modelling significance, e.g.: localised vertebrae are used as markers for detecting kyphosis or scoliosis, vertebral fractures, in surgical planning, or for follow-up analysis tasks such as  vertebral segmentation or their bio-mechanical modelling for load analysis. 

\noindent
\textbf{Vertebrae labelling.} Like several analysis approaches off-late, vertebrae labelling has seen successful utilisation of machine learning. One of the incipient and notable works by Glocker et al. \cite{glocker12}, followed by \cite{glocker13} used context-based features with regression forests and Markov models for labelling. In spite of their intuitive motivation, these approaches suffer a setback due to limited FOVs or presence of metal insertions. On a similar footing, \cite{suzani15} proposed a deep multi-layer perceptron using long-range context features. With the emergence of convolutional neural networks (CNN), Chen et al. \cite{chen15} proposed a joint-CNN as a combination of a random forest for initial candidate selection followed by a CNN trained to identify the vertebra based on its appearance and a conditional dependency on its neighbours. Without hand-crafting features this approach performed remarkably well. However, since the CNN works on a limited region around the vertebra, it results in a high variability of the localisation distance. Recently, Yang et al., with \cite{yang_ipmi} and \cite{yang_miccai}, proposed a deep, volumetric, fully-convolutional 3D network (FCN) called DI2IN with deep-supervision. The output of DI2IN is improved in subsequent stages that employ either message-passing across channels or a convolutional LSTM followed by further tuning with a shape dictionary.

Owing to equivariance of the convolutional operator and limited receptive field, an FCN doesn't always learn the anatomy of the region-of-interest. This is a severe limitation as human-equivalent learning utilises anatomical details aided with prior knowledge. An immediate remedy is to increase the receptive field by going deeper. However, this comes at the cost of higher model complexity or is just unfeasible due to memory constraints when working with volumetric data.

\noindent
\textbf{Prior \& adversarial learning in CNNs.} Recent work in \cite{ravishankar17} and \cite{oktay17} propose encoding (anatomical) segmentation priors into an FCN by learning the shape representation using an auto encoder (AE). The segmentation is expressed in terms of a pre-learnt latent space for evaluating a prior-oriented loss, which is then used to guide the FCN into predicting an anatomically sound segmentation. Our approach shares similarities with this approach with certain fundamental differences: (1) Our approach is aimed at localisation, which requires a redefinition of the notion of anatomical \emph{shape}. (2) We employ an AE for shape regularisation, but do not `pre-train' it to learn the latent space. We train the AE adversarially in tandem with the FCN. Parallels can be drawn between end-to-end learning of priors and learning the distribution of priors using generative adversarial networks (GANs). Both have two networks, a predictor (generator) and an auxiliary network which works on the `goodness' of the prediction. In medical image analysis where scan sizes are large and data are few, inspired from an energy-based adversarial generation framework (Zhao et al., \cite{ebgan16}), it is preferable to employ an adversary providing an anatomically-inspired supervision instead of the usual binary adversarial supervision (vanilla GAN).

\noindent
\textbf{Our contribution.} In this work, we propose an end-to-end solution for vertebrae labelling by adversarially training an FCN, thereby encoding the local spine structure into it. More precisely, relying on the sufficiency of information in certain 2D projections of 3D data, we propose: (1) A butterfly-shaped network that operates on 2D sagittal and coronal reformations, combining information across these views at a large receptive field, (2) Encoding the spine's structure into the Btrfly net using an energy-based, fully-convolutional, adversarial auto encoder acting as a discriminator. Our approach attains identification rates above 85\% without any post-processing stages, achieving state-of-art performance.

\begin{figure*}[t!]
 \centering
    \begin{subfigure}[t]{0.69\textwidth}
      \centering
      \hspace{-0.3in}
       \includegraphics[width=\textwidth]{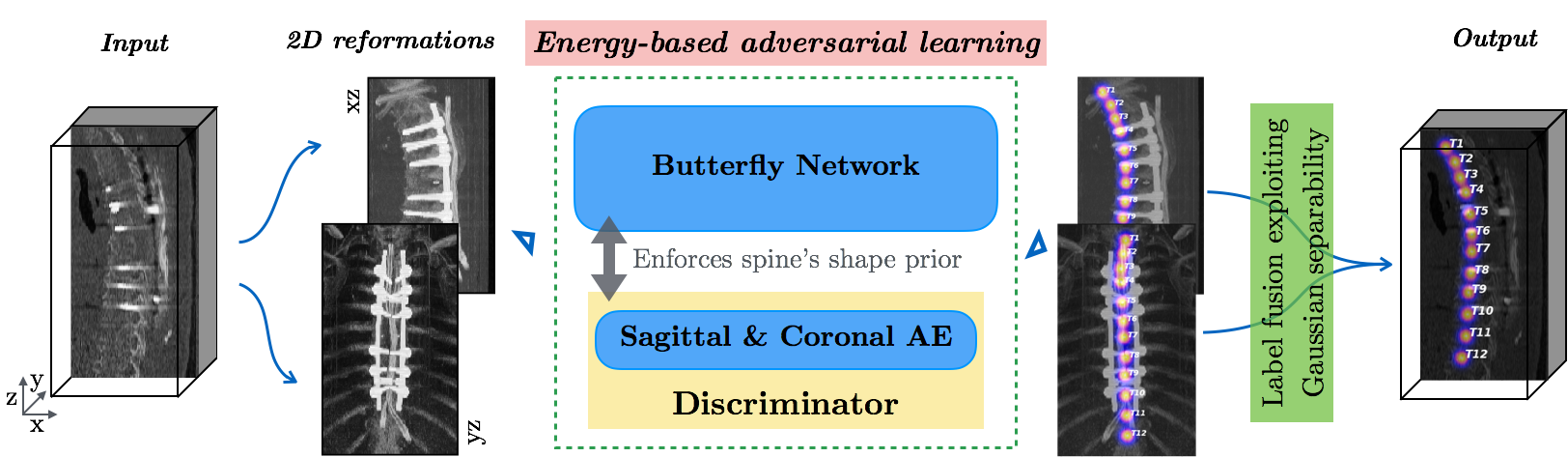}\\{(a)}
    \end{subfigure}
    ~
    \begin{subfigure}[t]{0.15\textwidth}
    \centering
       \includegraphics[width=\textwidth]{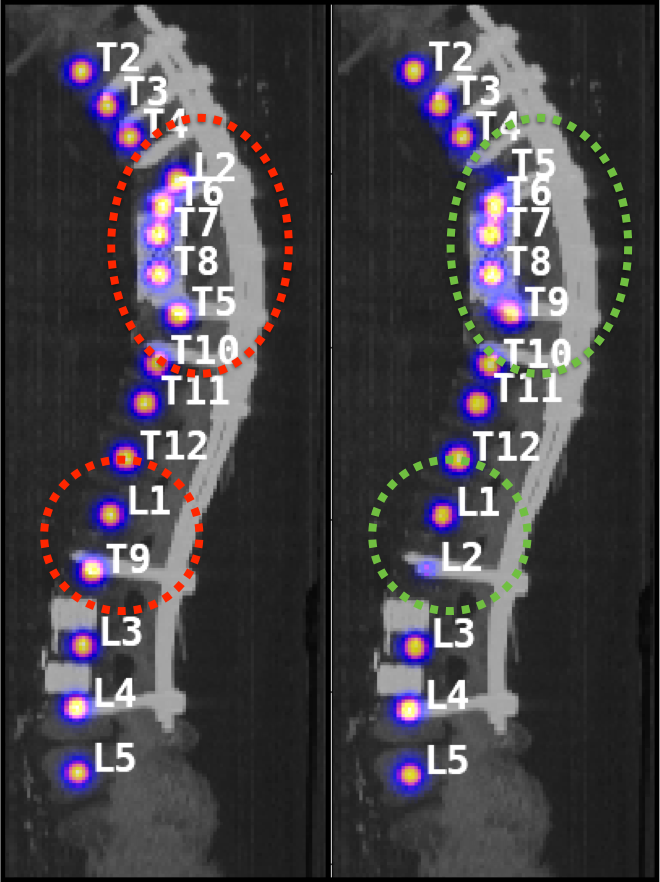}\\{(b)}
    \end{subfigure}
   \caption{\small (a) Overview of our approach. (b) Label correcting capability of the AE when trained as a denoising convolutional auto-encoder (red: corrupted, green: corrected). This motivates the discriminator in our adversarial framework.}
   \label{figure:1}
\end{figure*}

\vspace{-1mm}
\section{Methodology}
We present our approach in two stages. First, we describe the Btrfly network tasked with the labelling of the vertebrae. Then we present the adversarial learning of the local spine shape with an energy-based auto-encoder acting as the discriminator. Fig.~\ref{figure:1}a gives an overview of the proposed approach and the motivation for prior-encoding is illustrated in Fig.~\ref{figure:1}b.

\subsection{Btrfly Network}
\label{subsec:btrfly}
Working with 3D volumetric data is computationally restrictive, more so for localisation and identification that rely on a large context so as to capture spatially distant landmarks. Consequently, there is a trade-off between working with low-resolution data or resorting to shallow networks. Therefore, we propose working in 2D with \emph{sufficiently--representative} projections of the volumetric data. The choice of projection is application dependant. Since we are working with bone, we work on sagittal and coronal maximum intensity projections (MIP). The former captures the spine's curve and the latter captures the rib-vertebrae joints, both of which are crucial markers for labelling. Note that a naive MIP might not always be the optimal choice of projection, eg. in full-body scans where spine is not spatially centred or is obstructed by the ribcage in a MIP. Such cases are handled with a pre-processing stage detecting the occluded spine in the MIP (discussed in the Supplement, Sec. \ref{sec:suppl}).

\noindent
\textbf{Annotations.} We formulate the problem of learning the vertebrae labels as a multi-variate regression. The ground-truth annotation $\mathbf{Y} \in \mathbb{R}^{(h\times w \times d \times 25)}$ is a 25-channeled, 3D volume with each channel corresponding to each of the 24 vertebrae (C1 to L5), and one for the background. Each channel $i$ is constructed as a Gaussian heat map of the form $\mathbf{y}_{i} = e^{-{||x-\mu_i||^2}/{2\sigma^2}}$, $x \in \mathbb{R}^3$ where $\mu_i$ is the location of the $i^{th}$ vertebra and $\sigma$ controls the spread. The background channel is constructed as, $\mathbf{y}_0 = 1 - \max_{i}(\mathbf{y}_i)$. The sagittal and coronal MIPs of $\mathbf{Y}$ are denoted by $\mathbf{Y}_\text{sag}  \in \mathbb{R}^{(h\times w \times 25)}$ and  $\mathbf{Y}_\text{cor} \in \mathbb{R}^{(h\times d \times 25)}$, respectively.

\begin{figure*}[t]
 \centering
 	\includegraphics[width=0.8\textwidth]{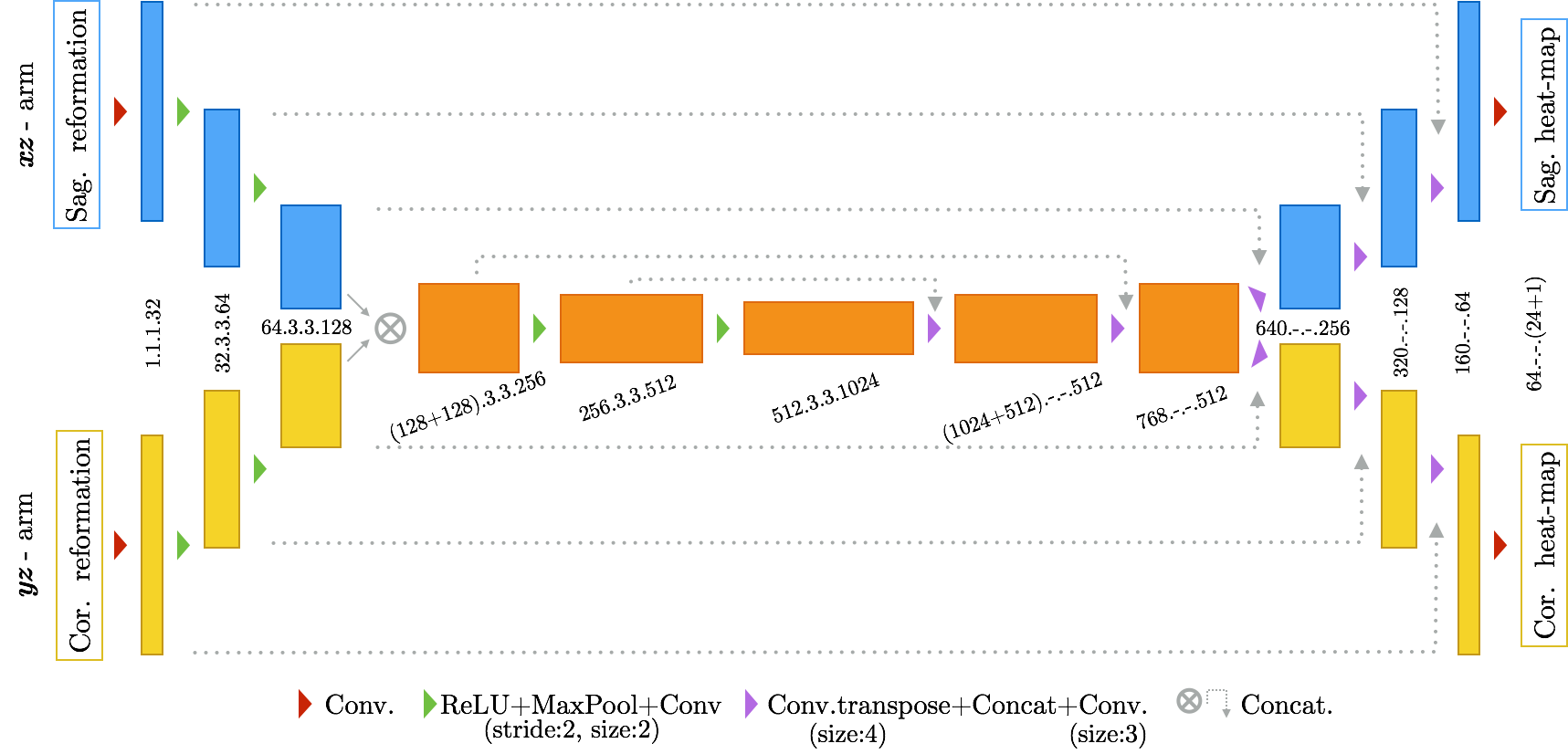}
         \caption{\small{The Btrfly architecture. The xz- (blue) and the yz-arms (yellow) correspond to the sagittal and coronal views. The kernel's shape resulting in each of the blocks is indicated as: \{input channels\} $\cdot$ \{kern. height\} $\cdot$ \{kern. width\} $\cdot$ \{output channels\}}}
   \label{figure:2}
\end{figure*}

\noindent
\textbf{Architecture.} We employ an FCN to perform the task of labelling. Since essential information is contained in both the sagittal and coronal reformations, and since the spine is approximately spatially centred in both, fusing this information across views leads to an improved identification. We propose a butterfly-like network (cf. Fig~\ref{figure:2}) with two arms (xz- and yz-arms) each concerned with one of the views. The feature maps of both the views are combined after a certain depth in order to learn their inter-dependency. 

\noindent
\textbf{Loss.} We choose an $\ell_2$ distance as the primary loss supported by a cross-entropy loss over the softmax excitation of the ground truth and the prediction. The total loss is expressed as:
\begin{equation}
\mathcal{L}_{\text{b,sag}} = ||\mathbf{Y}_\text{sag} - \tilde{\mathbf{Y}}_\text{sag}||^2+\omega H\mathbf{(Y}_\text{sag}^\sigma, \tilde{\mathbf{Y}}_\text{sag}^\sigma),
\label{eq:1}
\end{equation} 
where $\tilde{\mathbf{Y}}_\text{sag}$ is the prediction of the net's xz-arm, $H$ is the cross-entropy function, and $\mathbf{Y}_\text{sag}^\sigma = \sigma(\mathbf{Y}_\text{sag})$, the softmax excitation. $\omega$ is the median frequency weighing map (described in \cite{roy17}), boosting the learning of less frequent classes. The loss for the yz-arm is constructed in a similar fashion and the total loss of the Btrfly net is given by $\mathcal{L}_{\text{b}} = \mathcal{L}_{\text{b,sag}} + \mathcal{L}_{\text{b,cor}}$.

\subsection{Energy-based adversary for encoding prior}
\label{subsec:btrfly_pe}
Since the Btrfly net is fully-convolutional, its predictions across voxels are independent of each other owing to the spatial invariance of convolutions. Whatever information it encodes is solely due to its receptive field, which may not be anatomically consistent across the image. We propose to impose the anatomical prior of the spine's shape onto the Btrfly net with \emph{adversarial} learning.

\begin{figure*}[t!]
 \centering
    \begin{subfigure}[t]{0.4\textwidth}
       \centering
       $\vcenter{\includegraphics[width=\textwidth]{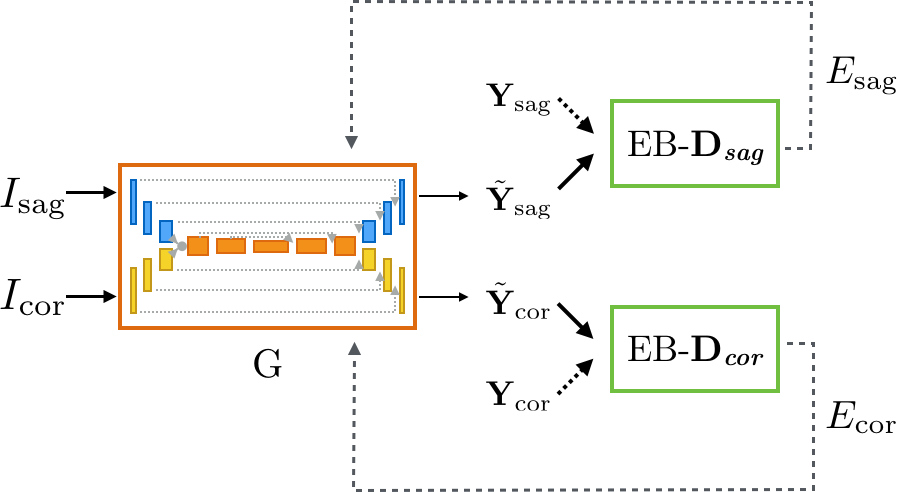}\vfill{(a)}}$
    \end{subfigure}
    ~
    \begin{subfigure}[t]{0.5\textwidth}
    \centering
       $\vcenter{\includegraphics[width=\textwidth]{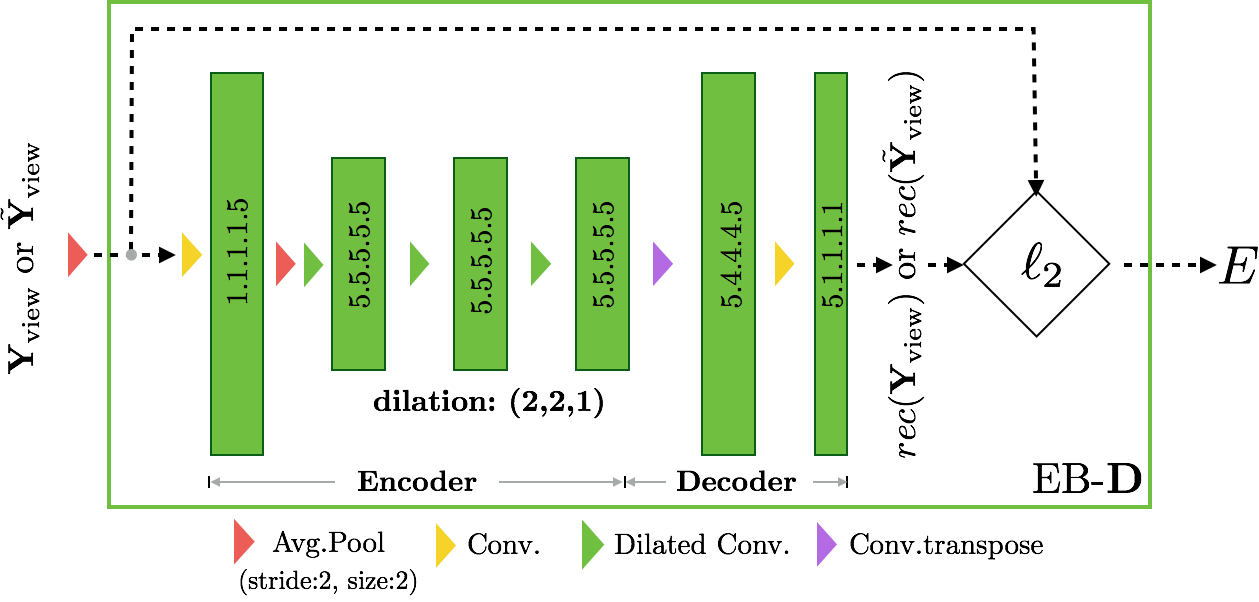}\vfill{(b)}}$
    \end{subfigure}
   \caption{\small(a) A overview of adversarial training showing the input to, and the energy-based supervision signal from, the discriminators. (b) The composition of the energy-based discriminator (EB-$D$). It gives the $\ell_2$ reconstruction error as output.}
      \label{figure:3}
\end{figure*}

Denoting the projected annotation as $\mathbf{Y}_\text{view}$, where view$\in$\{sag,cor\}, a sample annotation consists of a 2D Gaussian at the vertebral location in each channel (except $\mathbf{y}_0$). Looking at $\mathbf{Y}_\text{view}$ as a 3D volume enables us in learning the spread of Gaussians across channels and consequently the vertebral labels. However, owing to the extreme variability of FOVs and scan sizes, it is preferable to learn the spread of the vertebrae in parts. Therefore, we employ a fully-convolutional, 3D auto encoder (AE) with a receptive field covering a part of the spine at a time. The absence of  fully-connected layers in the AE also removes the necessity to resize the data, making it end-to-end trainable with the Btrfly net. Fig.~\ref{figure:3}a shows the arrangement of the AEs as adversaries w.r.t the Btrfly net. In an adversarial framework, the Btrfly net acts as the generator ($G$), and the local manifolds learnt from $\mathbf{Y}_\text{view}$ influence $\tilde{\mathbf{Y}}_\text{view}$ and vice versa.

\noindent
\textbf{Discriminator.} We devise the 3D adversary ($D$, cf. Fig.~\ref{figure:3}b) consisting of the AE as a functional predicting the $\ell_2$ distance between the input $\mathbf{Y}_\text{view}$ and its reconstruction by the AE, $rec(\mathbf{Y}_\text{view})$:  $D(\mathbf{Y}_\text{view})=E=||\mathbf{Y}_\text{view} - rec(\mathbf{Y}_\text{view})||^2$. This energy, $E$ is fed back into $G$ for adversarial supervision, as in \cite{ebgan16}. As it is an energy-based functional, we interchangeably refer to the discriminator as EB-$D$. Since $\mathbf{Y}_\text{view}$ consists of Gaussians, it is less informative than an image. Therefore, we avoid using max-pooling by resorting to average pooling. In order to have a receptive field covering multiple vertebrae without using pooling operations, we employ spatially dilated convolution kernels \cite{Yu2015MultiScaleCA} of size ($5\times5\times5$) with a dilation rate of 2 (only in image plane), resulting in a receptive field of $76\times76$ pixels. At \siunit{1}{mm} isotropic resolution, this covers 2 to 3 vertebrae in the lumbar region and more elsewhere.

\noindent
\textbf{Losses.} As in any adversarial setup, EB-$D$ is shown real ($\mathbf{Y}_x$($\equiv\mathbf{Y}_\text{view}$))  and generated annotations ($\mathbf{Y}_g$($\equiv\tilde{\mathbf{Y}}_\text{view}$)), and it learns to discriminate between both by predicting a low $E$ for real annotations, while $G$ learns to generate annotations that would trick $D$. For a given positive, scalar margin $m$, the following generator and discriminator losses are optimised:
\begin{equation}
\mathcal{L}_D = D(\mathbf{Y}_x) + \max(0,m - D(\mathbf{Y}_g)), \text{ and}
\label{eq:2}
\end{equation}
\begin{equation}
\mathcal{L}_G = D(\mathbf{Y}_g) + \mathcal{L}_\text{b,view}.
\label{eq:3}
\end{equation}
The joint optimisation of (\ref{eq:2}) and (\ref{eq:3}) for both the EB-$D$s results in a $G$ that performs vertebrae labelling while respecting the spatial distribution of the vertebrae across channels. We refer to this prior-encoded $G$ as the `Btrfly$_\text{pe}$' net.

\subsection{Inference}
\label{subsec:inference}
Once trained, an inference for a given input scan of size $(h\times w\times d)$ proceeds as: the desired sagittal and coronal MIP reformations are obtained and given as input to the xz- and yz-arms of the Btrfly net, resulting in a $(h \times w \times 25)$ sagittal heatmap and $(h \times d \times 25)$ coronal heatmap. The values below a threshold ($T$, selected on validation set) are ignored in order to remove noisy predictions. As the Gaussian kernel is separable, an outer product of the predictions results in the final heat map as
$
\tilde{\mathbf{Y}} = \tilde{\mathbf{Y}}_\text{sag} \otimes \tilde{\mathbf{Y}}_\text{cor},
$
where $\otimes$ denotes the outer product. The 3D location of the vertebral centroids are obtained as the maxima in their corresponding channels. Note that the EB-$D$ is no longer required during inference as its role in encoding the prior ends with the convergence of the Btrfly$_{\text{pe}}$ net.

\begin{figure*}[t!]
 \centering
 	\includegraphics[width=0.77\textwidth]{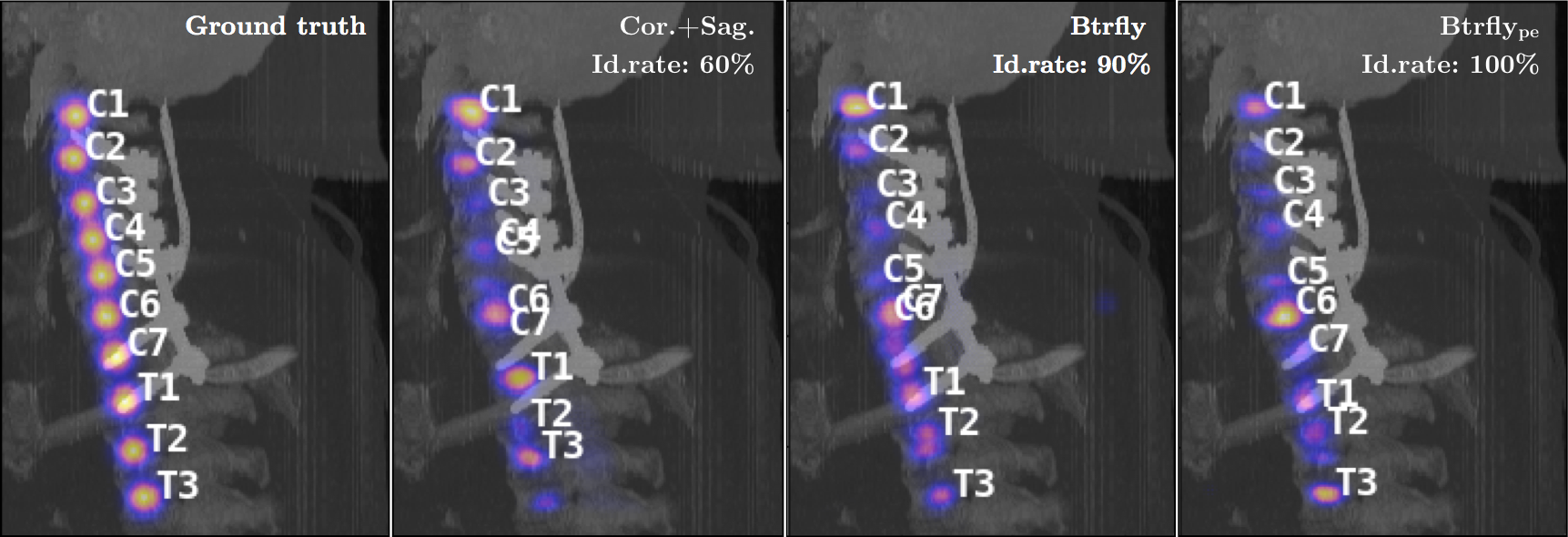}
         \caption{\small Effect of prior encoding: the prior-encoded Btrfly$_\text{pe}$ net successfully performs its task of prevent overlapping labels (C6 \& C7), consequently reordering all the vertebral labels. The reported id. rates are per volume. (Addn. results in Supplement, Sec. \ref{sec:suppl}.)}
   \label{figure:4}
\end{figure*}

\section{Experiments}
\label{sec:exp}
The evaluation is performed using a dataset introduced  in \cite{glocker13} with a total of 302 CT scans (242 for training and 60 for testing) including various challenges such as scoliotic spines, metal insertions, and highly restrictive FOVs. However, these are cropped to a region around the spine which excludes the ribcage. Thus, a naive sagittal and coronal MIP, without any pre-processing, suffices to obtain the input images for our approach. In order to enhance the net's robustness, 10 MIPs are obtained from one 3D scan, each time randomly choosing half the slices of interest. This leads to a total of 2420 reformations per view for training (incl. a validation split of 100). We present the experiments with the Btrfly net trained as stand-alone as well as with the prior-encoding discriminator EB-D. Batch-normalisation is used after every convolution layer, along with 20\% dropout in the fused layers of Btrfly. Additionally, so as to validate the necessity of the combination of views, we compare the Btrfly net's performance with that of two networks working individually on the views (denoted as Cor.+Sag. nets). The architecture of each of these networks is similar to one arm of the Btrfly net. The optimiser's setup in all the three cases is similar: an Adam optimiser is employed with an initial learning rate of $\lambda=1\times 10^{-3}$, working on data resampled to a \siunit{1}{mm} isotropic resolution. $\lambda$ is decayed by a factor of $3/4th$ every 10k iterations to $0.2\times 10^{-3}$. Convergence of all the networks is tested on the validation set.

\begin{table*}[h]
\small
 \newcommand{\tabincell}[2]{\begin{tabular}{@{}#1@{}}#2\end{tabular}}
 \renewcommand\arraystretch{1}
 \centering
 \setlength{\tabcolsep}{0.4em}
 \caption{\small Performance comparison of our approach (setting $T=0$, for a fair comparison) with Glocker et al. \cite{glocker13}, Chen et al. \cite{chen15}, \& Yang et al. \cite{yang_ipmi}. DI2IN refers to stand-alone FCN, while DI2IN* includes use of message passing and shape dictionary. We do not compare with experiments  in \cite{yang_ipmi} that use additional undisclosed data.}

\begin{tabular}{ c | c : c : c : c | c : c : c}
\specialrule{.1em}{0em}{-.1em}
 \rule{0pt}{2.5ex}Measures&\cite{glocker13}&\cite{chen15}&DI2IN\cite{yang_ipmi}&DI2IN$^*$\cite{yang_ipmi}&Cor.+Sag.&Btrfly&\textbf{Btrfly$_\text{pe}$}\\ [0.25ex]
\specialrule{.05em}{-0.1em}{0em}
 \rule{0pt}{2.5ex}Id.rate & 74.0 & 84.2 & 76.0 & 85.0 & 78.1 & 81.8 & \textbf{86.1}\\
 d$_\text{mean}$& 13.2 & 8.8 & 13.6 & 8.6 & 9.3 & 7.5 & \textbf{7.4} \\
 d$_\text{std}$ & 17.8 & 13.0 & 37.5 & 7.8 & 8.0 & \textbf{5.4} & 9.3\\
 \specialrule{.1em}{0em}{0em}
\end{tabular}
\label{table:1}
\end{table*}

\begin{figure}[t]
\begin{floatrow}
\ffigbox{%
  \includegraphics[width=0.5\textwidth]{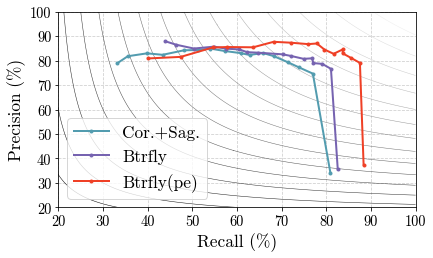}
}{%
  \caption{\small A precision-recall curve with F1 isolines, illustrating the effect of the $T$ during inference. For any $T$, Btrfly$_\text{pe}$ offers a better trade-off between $P$ and $R$.
  \label{figure:5}
  }%
}
\hfill
\capbtabbox[0.48\textwidth]{%
\small
       \begin{tabular}{cccc}
 	\toprule
	\rule{0pt}{2ex}Approach&$P$&$R$&F1\\ [0.25ex] 
 	\midrule
	 \rule{0pt}{2.5ex}Cor.+Sag.$_\text{($T$=0.05)}$ & 74.7 & 77.0 & 75.8 \\
	\rule{0pt}{2ex}Btrfly$_\text{($T$=0.1)}$ & 78.7 & 79.1 & 78.9 \\
	\rule{0pt}{2ex}\textbf{Btrfly}$_\text{\textbf{pe} ($T$=0.2)}$ & \textbf{84.6} & \textbf{83.7} & \textbf{84.1}\\
 	\midrule
	\end{tabular}
	\vspace{0.35in} 
}
{%
  \caption{\small The optimal $P$ and $R$ values based on F1 score, along with the optimal $T$. $R$ at optimal-F1 of Btrfly$_\text{pe}$ is comparable to state-of-art.}
  \label{table:2}%
}
\end{floatrow}
\end{figure}

\noindent
\textbf{Evaluation \& discussion.} For evaluating the performance of our network with prior work, we use two metrics defined in \cite{glocker12} namely, the \emph{identification rates} (id. rate, in \%) and \emph{localisation distances} (d$_\text{mean}$ \& d$_\text{std}$, in $\mathrm{mm}$). We report the measures in Table \ref{table:1}. It lists the performance of three variants of our network and compares them with several recent approaches. We address three main questions through our experiments: \emph{(1) Why the butterfly shape?} Compared to Cor.+Sag. nets, performance  improves with the Btrfly net. This is because the combination of views causes the predictions of the Btrfly net to be spatially consistent across views. We also observe a 6\% improvement in the id.rate over a naive 3D FCN (DI2IN). \emph{(2) Why the adversarial prior-encoding?} In addition to the advantages of the Btrfly net, the Btrfly$_\text{pe}$ net possesses adversarially encoded spatial distribution of the vertebrae. This results in about a 4\% increase in the id. rate. Compared to the prior work, Btrfly$_\text{pe}$ net achieves state-of-art measures in both the metrics, and it does so by being a single network trained end-to-end. (cf. Fig.~\ref{figure:4})  \emph{(3) Relation to latent-space learning?} EB-$D$ is more flexible than the AEs in \cite{oktay17,ravishankar17} as it learns from scratch and converges to a latent manifold best representing the true as well as generated data. The reconstruction capability of the AE for a generated sample is of interest. Using the output of the AE instead of Btrfly$_\text{pe}$, we achieve an id.rate of 75\% with a d$_\text{mean}$ of \siunit{19}{mm}, indicating the AEs' capability of transferring the learning from true to contrastive samples. 

\noindent
\textbf{Precision \& Recall.} Localisation distance and id.rate capture the ability of the network in accurately labelling a vertebra. However, both the measures are agnostic to false positive predictions. Accounting for spurious predictions becomes important especially when dealing with FCNs, as the predictions depend on a locally constrained receptive field. In our case, the false positives are controlled by the threshold $T$ as described in Section \ref{subsec:inference}. Accounting for these, we define two measures, \emph{precision} ($P$) and \emph{recall} ($R$) as: 
$
P = \nicefrac{\#\text{hits}}{\#\text{predicted}}
$ 
and
$
R = \nicefrac{\#\text{hits}}{\#\text{actual}}
$, where $\#$hits is the number of vertebrae satisfying the condition of identification as defined for id.rate, $\#$predicted is the vertebrae in the prediction, and $\#$actual is the vertebrae actually present in the image. Observe that id. rate is measured over all vertebrae in the test set while $R$ is measured \emph{per scan} and averaged over test scans. Fig.~\ref{figure:5} shows a precision-recall curve generated by varying $T$ between 0 to 0.8 in steps of 0.05, while Table~\ref{table:2} shows the performance at the F1-optimal threshold. In spite of not choosing an recall-optimistic threshold, our networks perform comparably well. Notice the over-arcing nature of Btrfly over Cor.+Sag. nets and that of Btrfly$_\text{pe}$ over others.

\section{Conclusions}
We validate the sufficiency of 2D orthogonal projections of the spine for localising and identifying the vertebrae by combining information across the projections using a butterfly-like architecture. In addition to looking at a local receptive field like any FCN, our approach considers the local structure of the spine thanks to an adversarial energy-based prior encoding, thereby outperforming the state-of-art approaches as a stand-alone network without any post-processing stages.\\

\noindent
\textbf{Acknowledgements.} This work is supported by the European Research Council (ERC) under the European Union's `Horizon 2020' research \& innovation programme (GA637164--iBack--ERC--2014--STG). We acknowledge NVIDIA Corporation's support with the donation of the Quadro P5000 used for this research.

\bibliographystyle{splncs03}
\bibliography{bibliography}

\begin{thebibliography}{10}
\providecommand{\url}[1]{\texttt{#1}}
\providecommand{\urlprefix}{URL }

\bibitem{chen15}
Chen, H., et~al.: Automatic localization and identification of vertebrae in
  spine ct via a joint learning model with deep neural networks. In: MICCAI.
  pp. 515--522. Springer (2015)

\bibitem{glocker12}
Glocker, B., et~al.: Automatic localization and identification of vertebrae in
  arbitrary field-of-view ct scans. In: MICCAI. pp. 590--598. Springer (2012)

\bibitem{glocker13}
Glocker, B., et~al.: Vertebrae localization in pathological spine ct via dense
  classification from sparse annotations. In: MICCAI. pp. 262--270. Springer
  (2013)

\bibitem{oktay17}
Oktay, O., et~al.: Anatomically constrained neural networks {(ACNN):}
  application to cardiac image enhancement and segmentation. CoRR
  abs/1705.08302 (2017)

\bibitem{ravishankar17}
Ravishankar, H., et~al.: Learning and incorporating shape models for semantic
  segmentation. In: MICCAI. pp. 203--211. Springer (2017)

\bibitem{roy17}
Roy, A.G., et~al.: Error corrective boosting for learning fully convolutional
  networks with limited data. In: MICCAI. pp. 231--239. Springer (2017)

\bibitem{suzani15}
Suzani, A., et~al.: Fast automatic vertebrae detection and localization in
  pathological ct scans - a deep learning approach. In: MICCAI. pp. 678--686
  (2015)

\bibitem{yang_ipmi}
Yang, D., et~al.: Automatic vertebra labeling in large-scale 3d ct using deep
  image-to-image network with message passing and sparsity regularization. In:
  IPMI. pp. 633--644. Springer (2017)

\bibitem{yang_miccai}
Yang, D., et~al.: Deep image-to-image recurrent network with shape basis
  learning for automatic vertebra labeling in large-scale 3d ct volumes. In:
  MICCAI. pp. 498--506. Springer (2017)

\bibitem{Yu2015MultiScaleCA}
Yu, F., Koltun, V.: Multi-scale context aggregation by dilated convolutions.
  In: ICLR (2016)

\bibitem{ebgan16}
Zhao, J.J., et~al.: Energy-based generative adversarial network. CoRR
  abs/1609.03126 (2016)

\end{thebibliography}


\begin{thebibliography}{5}
\bibitem{ssd}
Liu W. et al.: SSD: Single Shot MultiBox Detector. CoRR abs/1512.02325 (2015), \\\url{http://arxiv.org/abs/1512.02325}.
\bibitem{xvertseg} 
The xVertSeg Challenge, \url{http://lit.fe.uni-lj.si/xVertSeg/}
\end{thebibliography}

\newpage
\newgeometry{left=20mm,right=20mm, top=20mm, bottom=20mm}

\section{Supplementary Material}
\label{sec:suppl}
\subsection{Case study on a non-spine-centred scan} 
The benchmark dataset used in Section 3 of our work is mostly spine-centred, and the naive maximum intensity projections contain no occlusions. However, in certain full-body scans, the spine is obstructed by the ribcage in a MIP of the entire scan, or the spine is not spatially centred in both the views, thus not taking full advantage of Btrfly net's view fusion (cf. Fig.~\ref{figure:suppl}a). Such cases can be handled by a introducing a pre-processing step before the Btrfly$_\text{pe}$ net in the form of an `object-detection' network. 

For such scenario, we construct the MIPs in two stages. The first MIP is constructed on the entire scan. On this, we use a \emph{single-shot object detection} (SSD) inspired architecture \cite{ssd}  trained to identify occluded spines (cf. Fig.~\ref{figure:suppl}a). Once the spine is located,  we construct the second pair of MIPs based on the \emph{spine}-slices, which are then used as inputs to the Btrfly$_\text{pe}$ net (cf. Fig.~\ref{figure:suppl}b,c). The ground truth for the SSD net can be constructed from the ground truth annotation of the vertebral centroids. We use a generic 16-layer residual CNN with an SSD extension. This use-case  is illustrated on a scan from the training set of the xVertSeg \cite{xvertseg} dataset. Note that we used the xVertSeg data only for inference and not for re-training the network. The centroids of the vertebrae are obtained from the maximum point of the distance transform of the segmentation map (xVertSeg has voxel-level annotations from L1 to L5).

\setcounter{figure}{5}
\begin{figure*}
\centering
   
   \begin{subfigure}[b]{0.46\textwidth}
   \includegraphics[width=1\linewidth]{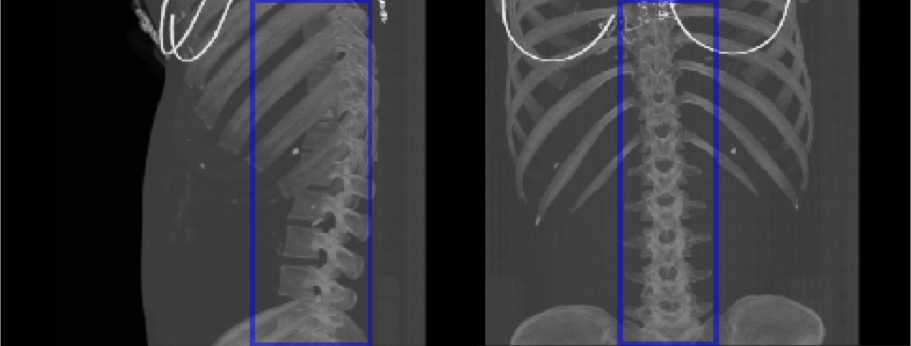}
   \caption{}
   \end{subfigure}
~
   \begin{subfigure}[b]{0.47\textwidth}
   \includegraphics[width=1\linewidth]{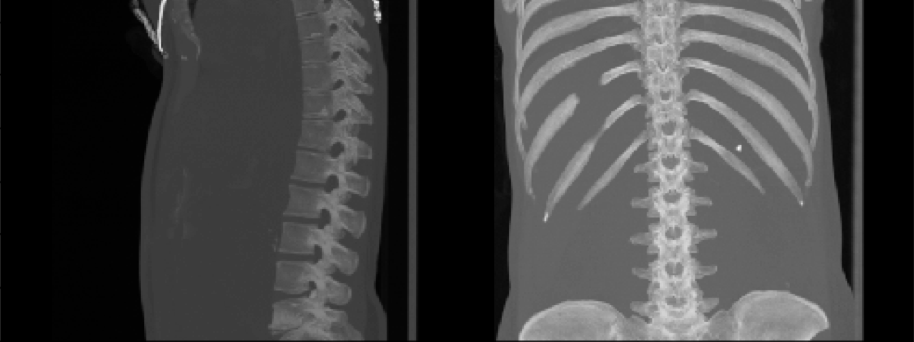}
   \caption{}
   \end{subfigure}
   
   \begin{subfigure}[b]{0.47\textwidth}
   \includegraphics[width=1\linewidth]{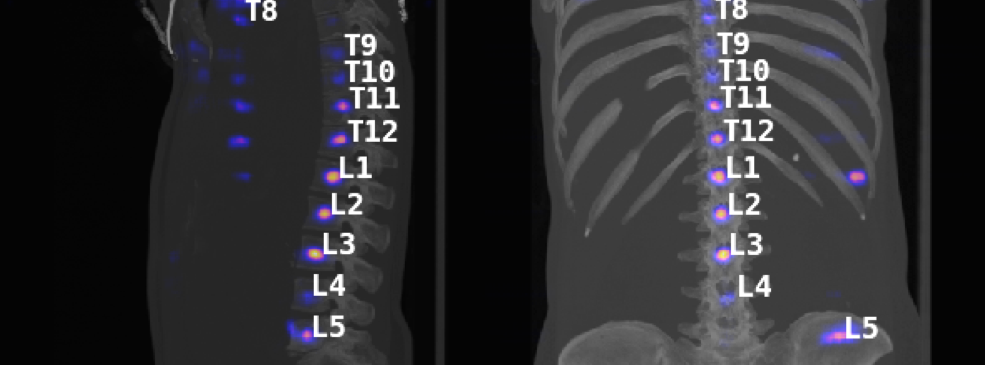}
   \caption{}
   \end{subfigure}
   ~
   \begin{subfigure}[b]{0.47\textwidth}
   \includegraphics[width=1\linewidth]{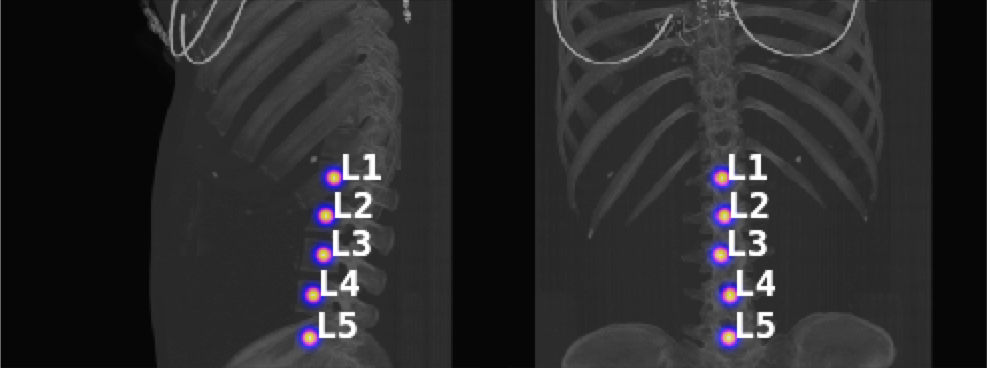}
   \caption{}
   \end{subfigure}
\caption{\small{An illustration of the extension to Brtfly$_\text{pe}$ net. \textbf{(a)} Naive sagittal and coronal MIPs on the entire scan with the bounding box predictions (in blue) of our SSD net. Observe the ribcage obstructing the spine. \textbf{(b)} Improved  MIPs constructed from the slices containing the spine based on the localisation in (a). \textbf{(c)} Output of the Btrfly$_\text{pe}$ net, resulting in an 80 \% id.rate. Also observe the incorrect localisation of T8 and L5, along with prediction noise in sagittal view owing to the non-aligned spine in both views. We believe that aligning the spine using its detection could further improve the prediction. \textbf{(d)} The ground truth centroids constructed from the voxel-level annotation map of scan. Since xVertSeg data only has lumbar annotations, we visualise the lumbar centroids.}}
\label{figure:suppl}
\end{figure*}

\begin{figure*}
\centering
   \includegraphics[width=0.8\linewidth]{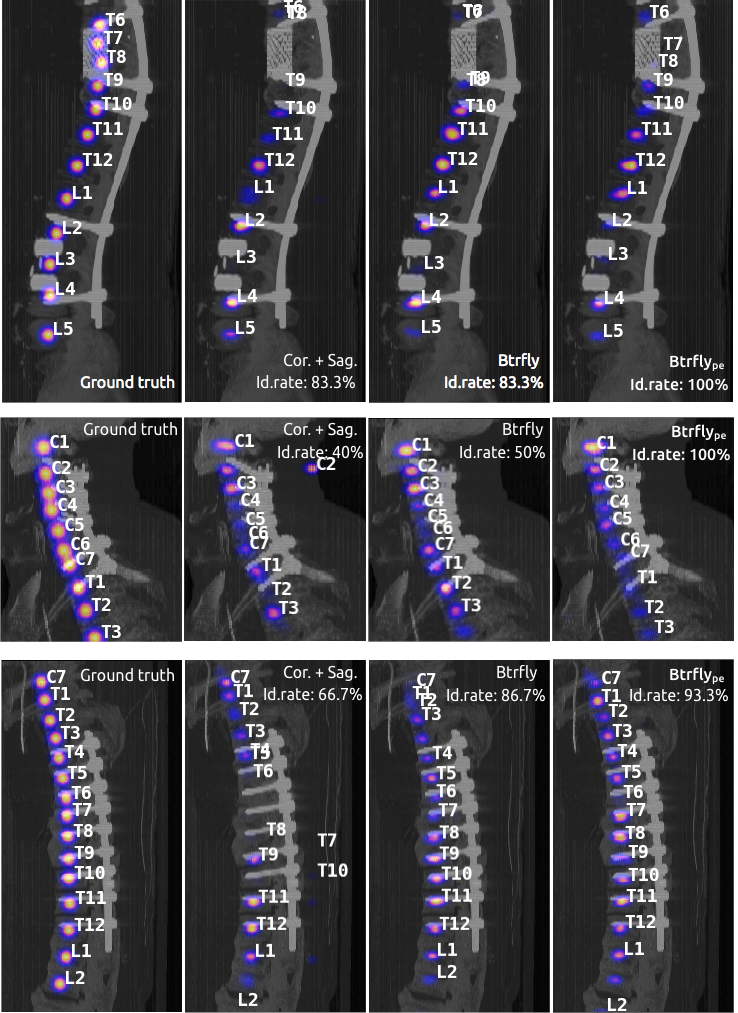}
\caption{\small{\textbf{Additional quantitative results.} MIP images with predictions of the three variants of our approach at $T$=0 for all cases. The spine's local structure is conserved in the predictions of Btrfly$_{pe}$. Also observe that, as a consequence of prior encoding, in some cases labels are predicted in spite of no useful spatial information, albeit the strength of these predictions is less.}}
\label{figure:qual}
\end{figure*}

\end{document}